\documentclass{article}

\usepackage{arxiv}
\usepackage{booktabs} 
\usepackage[utf8]{inputenc}
\usepackage{subcaption}
\usepackage{csquotes}
\usepackage{xcolor}
\usepackage{graphicx}
\usepackage{amsmath}
\usepackage{algorithm}
\usepackage[noend]{algorithmic}
\usepackage{float}
\usepackage{soul}
\usepackage{longtable}
\usepackage{multirow}
\usepackage{tikz}
\usepackage{pgfplots}
\sethlcolor{lightgray}

\pgfplotsset{compat=1.17}
\title{Computing High-Quality Solutions for the Patient Admission Scheduling Problem using Evolutionary Diversity Optimisation}
\author{ Adel Nikfarjam \\
Optimisation and Logistics\\School of Computer Science\\The University of Adelaide\\
  \texttt{adel.nikfarjam@adelaide.edu.au} \\
  \And
  Amirhossein Moosavi\\ Health Systems\\
Telfer School of Management\\University of Ottawa\\
  \texttt{smoos018@uottawa.ca} \\
  \And
  Aneta Neumann\\
Optimisation and Logistics\\School of Computer Science\\The University of Adelaide\\
  \texttt{aneta.neumann@adelaide.edu.au} \\
  \And
   Frank Neumann \\
Optimisation and Logistics\\School of Computer Science\\The University of Adelaide\\
  \texttt{frank.neumann@adelaide.edu.au} \\}

\begin{document}
%
%
%
\maketitle              
\begin{abstract}
Diversification in a set of solutions has become a hot research topic in the evolutionary computation community. It has been proven beneficial for optimisation problems in several ways, such as computing a diverse set of high-quality solutions and obtaining robustness against imperfect modeling. For the first time in the literature, we adapt the evolutionary diversity optimisation for a real-world combinatorial problem, namely patient admission scheduling. We introduce an evolutionary algorithm to achieve structural diversity in a set of solutions subjected to the quality of each solution. We also introduce a mutation operator biased towards diversity maximisation. Finally, we demonstrate the importance of diversity for the aforementioned problem through a simulation.    
\end{abstract}
%
%
%

\section{Introduction}\label{sec1}


Traditionally, researchers seek a single (near) optimal solution for a given optimisation problem. Computing a diverse set of high-quality solutions is gaining increasing attention in the evolutionary computation community. Most studies consider diversity as finding niches in either fitness landscape or a predefined-feature space. In contrast, a recently introduced paradigm, \emph{evolutionary diversity optimisation}~(EDO), explicitly maximises the structural diversity of a set of solutions subjected to constraints on their quality. EDO has shown to be beneficial in several aspects, such as creating robustness against imperfect modeling and minor changes in problems' features \cite{NikfarjamB0N21b}. This paradigm was first defined by \cite{ulrich2011maximizing}. Afterwards, \cite{alexander2017evolution,doi:10.1162/evcoa00274} investigated the use of EDO in generating images and \emph{traveling salesperson problem} (TSP) benchmark instances, respectively. \cite{neumann2018discrepancy,neumann2019evolutionary} extended the previous studies and incorporated the concepts of star discrepancy and indicators from the frameworks of multi-objective evolutionary algorithms in EDO. More recently, diverse TSP solutions are evolved, using distance-based and entropy measure in \cite{viet2020evolving,NikfarjamBN021a}. A special EAX-based crossover that focuses explicitly on the diversity of TSP solutions is introduced by \cite{NikfarjamB0N21b}. Several studies examine EDO in combinatorial problems, such as the quadratic assignment problem~\cite{DoGN021}, the minimum spanning tree problem~\cite{Bossek021tree}, the knapsack problem~\cite{BossekN021KP}, the optimisation of monotone sub-modular functions~\cite{NeumannB021}, and the traveling thief problem \cite{NikTTPEDO}. Neumann rt al.~\cite{AnetaCo-EA} introduced a co-evolutionary algorithm to compute Pareto-front for bi-objective optimisation problems and concurrently evolve another set of solutions to maximise structural diversity. For the first time, we incorporate EDO into a real-world combinatorial problem, namely the \emph{patient admission scheduling}~(PAS). PAS is a complex multi-component optimisation scheduling problem in healthcare, involving more features compared to the problems already studied in the literature of EDO.

A recent report on the global health spending of 190 countries shows that healthcare expenditure has continually increased and reached around US\$ 10 trillion (or 10\% of global GDP) \cite{WHO3}. Due to the ever-increasing demand and healthcare expenditures, there is a great deal of pressure on healthcare providers to increase their service quality and accessibility. Among the several obstacles involved in healthcare resource planning, the PAS problem is of particular significance, impacting organisational decisions at all decision levels \cite{Bast}. PAS has been studied under different settings, but it generally investigates the allocation of patients to beds such that both treatment effectiveness and patients' comfort are maximised. A benchmark PAS problem is defined in \cite{Deme1}, and an online database including 13 benchmark test instances, their best solutions and a solution validator are maintained in \cite{Deme2}. Various optimization algorithms have been proposed for the benchmark PAS problem, such as simulated annealing \cite{Cesc}, tabu search \cite{Deme1}, mixed-integer programming \cite{Bast11}, model-based heuristic \cite{Turh11}, and column generation \cite{Rang11}. The simulated annealing in \cite{Cesc} has demonstrated the best overall performance amongst the \emph{evolutionary algorithms}~(EAs) algorithms for the PAS problem.
The previous studies aimed for a single high-quality solution for the PAS problem. In contrast, our goal is to diversify a set of PAS solutions, all with desirable quality but some different structural properties. \par



Having a diverse set of solutions can be beneficial for PAS from different perspectives. PAS is a multi-stakeholder problem and requires an intelligent trade-off between their interests, a challenging issue because (i) stakeholders have conflicting interests, (ii) health departments have diverse admission policies and/or requirements, and (iii) decision-makers (or even health systems) have different values so they could have different decision strategies. A diverse set of high-quality solutions provides stakeholders with different options to reach an agreement. Moreover, most hospitals use manual scheduling methods \cite{duran2017solving}. Several subtle features are involved in such problems that are not general enough to be considered in the modelling but occasionally affect the feasibility of the optimal solutions (imperfect modelling). Once again, computing a diverse set of solutions can be highly beneficial in providing decision-makers with robustness against imperfect modelling and small changes in problems. Thus, the use of EDO can be a step forward in seeing applications of optimisation methods, particularly EAs, in the practice. In this study, we first define an entropy-based measure to quantify the structural diversity of PAS solutions. Having proposed an EA maximising diversity, we introduce three variants of a random neighbourhood search operators for the EA. The first variant is a change mutation with fixed hyper-parameters, the second variant uses a self-adaptive method to adjust its hyper-parameters during the search process, the last variant is a biased mutation to boost its efficiency in maximising diversity of PAS solutions. Finally, we conduct a comprehensive experimental investigation to (i) examine the performance of the EA and its operators, (ii) illustrate the effects of EDO in creating robustness against imperfect modelling.

We structure the remainder of the paper as follows. The PAS problem and diversity in this problem are formally defined in Section \ref{sec2}. The EA is introduced in Section~\ref{sec3}. Comprehensive experimental investigations are conducted in Section \ref{sec4}. Finally, Section \ref{sec5} presents conclusions and some remarks.

\section{Patient Admission Scheduling}\label{sec2}
\noindent
This section presents the definition of the benchmark PAS problem \cite{Deme1} to make our paper self-contained.\par

\section*{Problem definition}
\noindent
The PAS problem assumes that each patient has a known gender, age, \emph{Length of Stay} (LoS), specialty and room feature/capacity requirements/preferences. Similarly, the set of rooms is known in advance, each associated with a department, gender and age policies, and medical equipment (or features). The problem assumptions for the PAS can be further explained as follows:

\begin{itemize}
    \item Room and department: Room is the resource that patients require during their treatment ($r \in \mathcal{R}$). Each room belongs to a unique department.
    \item Planning horizon: The problem includes a number of days where patients must be allocated to a room during their course of treatment ($t \in \mathcal{T}$).
    \item LoS: Each patient ($p \in \mathcal{P}$) has fixed admission and discharge dates, by which we can specify her LoS ($t \in \mathcal{T}_p$).
    \item Room capacity (A1): Each room has a fixed and known capacity on each day ($CP_r$).
    \item Gender policy (A2): The gender of each patient is known (either female or male). Each room has a gender policy. There are four different policies for rooms $\{D,F,M,N\}$: (i) rooms with policy $F$ ($M$) can only accommodate female (male) patients, (ii) rooms with policy $D$ can accommodate both genders, but they should include only one gender on a single day, and (iii) rooms with policy $N$ can accommodate both genders at the same time. There will be a penalty if cases (i) and (ii) are violated. While $CG^{1}_{pr}$ measures the cost of assigning patient $p$ to room $r$ regarding the gender policy for room types $F$ and $M$, $CG^{2}$ specifies the cost of violating the gender policy for room type $D$. 
    \item Age policy (A3): Each department is specialised in patients with a specific age range. Thus, patients should be allocated to the rooms of a department that respect their age policy. Otherwise, there will be a penalty. $CA_{pr}$ determines the cost of assigning patient $p$ to room $r$ regarding the age policy.
    \item Department specialty (A4): Patients should be allocated to the rooms of a department with an appropriate specialty level. A penalty occurs if this assumption is violated. $CD_{pr}$ specifies the cost of assigning patient $p$ to room $r$ regarding the department speciality level. 
    \item Room specialty (A5): Like departments, patients should be allocated to rooms with an appropriate specialty level. $CB_{pr}$ defines the penalty of assigning patient $p$ to room $r$ regarding the room speciality level.
    \item Room features (A6): Patients may need and/or desire some room features. A penalty occurs if the room features are not respected for a patient. Note that the penalty is greater for the required features compared to the desired ones. $CF_{pr}$ shows the combined cost of assigning patient $p$ to room $r$ regarding the room features.
    \item Room capacity preference (A7): Patients should be allocated to rooms with either preferred or smaller capacity. $CR_{pr}$ specifies the cost of assigning patient $p$ to room $r$ regarding the room capacity preference.
    \item Patient transfer: There exists a fixed penalty for each time patients are transferred during their LoS, which is shown by $CT$.
\end{itemize}

Assumption A1 must always be adhered (a hard constraint), while the rest of the assumptions (A2-A7) can be violated with a penalty (soft constraints). \cite{Deme2} provides complementary information on the calculation of these penalties. The objective function of the PAS problem minimizes the costs of violating assumptions A2-A7 and patient transfers between rooms within their LoS.\par

\begin{figure}
    \centering
    \includegraphics{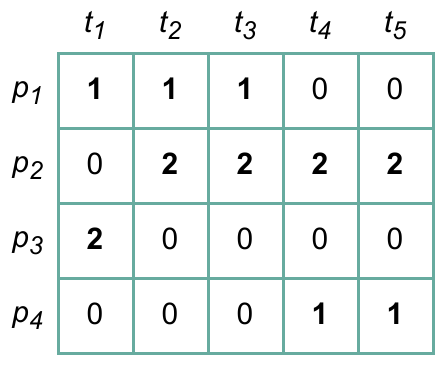}
    \caption{An illustrative example of the solution representation}
    \label{fig:solrep}
\end{figure}

To simplify the cost coefficients of the PAS problem, we merge the costs associated with assumptions A2-A7 (except the penalty of violating the gender policy for room type $D$) into a single matrix $CV_{pr}$ ($CV_{pr} = CA_{pr} + CD_{pr} + CB_{pr} + CF_{pr} + CR_{pr} + CG^{1}_{pr}$). Then, we formulate the objective function of this problem as below:

\begin{equation*} \label{eq1}
    \text{Min}_{s \in \Xi} \: O(s) = O^{1}(s, CV_{pr}) + O^{2}(s, CG^{2}) + O^{3}(s, CT)
\end{equation*}

\noindent
where $\Xi$ denotes the feasible solution space. This objective function minimises: (i) costs of violations from assumptions A2 (for rooms with policies $F$ and $M$) and A3-A7, (ii) costs of violations from assumption A2 (for rooms with policy $D$), and (iii) costs of patient transfers. We use a two-dimensional integer solution representation in this paper. Figure \ref{fig:solrep} illustrate an example of the solution representation with four patients, two rooms (each with one capacity), and a planning horizon of five days. Values within the figure represents the room number allocated to each patient. With this solution representation, we can compute Objective function \eqref{eq1} and ensure the feasibility of solutions with respect to the room capacity (the number of times a room appears in each column - each day - must be less than or equal to its capacity). For a mixed-integer linear programming formulation of this problem, interested readers are referred to \cite{Deme1}.\par

This study aims to maximise diversity in a set of feasible PAS solutions subjected to a quality constraint. Let (i) $S$ refers to a set of PAS solutions, (ii) $H(S)$ denotes diversity of population $S$, and (iii) $c_{max}$ refers to the maximum acceptable cost of solution $s$. Then, we define the diversity-based optimization problem as follows:

\begin{align*}
    &\max \: H(S) &\\
    &\text{subjected to:}&\\
    &O(s) \leq c_{max} & \forall s \in S
\end{align*}

Align with most studies in the literature of EDO, such as \cite{viet2020evolving,NikfarjamBN021a}, we assume that optimal/near-optimal solutions for given PAS instances are prior knowledge. Therefore, we set $c_{max} = (1+\alpha)O^*$, where $O^*$ is the optimal/near-optimal value of $O(s)$, and $\alpha$ is an acceptable threshold for $O^*$.\par

\subsection{Diversity in Patient Admission Scheduling}
In this sub-section, we define the diversity of PAS solutions, for which we require a measure quantifying the difference between solutions. We adopt a measure based on the concept of entropy. Here, entropy is defined on the number of solutions that patient $p$ is assigned to room $r$ on day $t$, $n_{prt}$. The entropy of population $S$ can be calculated from:

\begin{align*}
H(S) = \sum_{p \in \mathcal{P}} \sum_{r \in \mathcal{R}} \sum_{t \in \mathcal{T}_p} h(n_{prt}) \text{ with } h(x)=-\left(\frac{x}{\mu}\right)\ln{\left(\frac{x}{\mu}\right)}.
\end{align*}
where, $h(n_{prt})$ is the contribution of $n_{prt}$ to entropy of $S$. Note that $h(n_{prt})$ is set to zero if $n_{prt} = 0$. We, now, calculate the maximum achievable entropy. If $\mu \leq|\mathcal{R}|$, the maximum entropy ($H_{max}$) occurs when patient $p$ is not assigned to the same room $r$ at any days in $S$. In other words, $n_{prt}$ is at most equal to $1$. Generally, $H \gets H_{max} \iff max(n_{prt})=\lceil\mu/|\mathcal{R}|\rceil$. Based on the pigeon holds principle, the maximum entropy can be calculated from:
$$H_{max} = W(\mu\mod|\mathcal{R}|)h(\lceil\mu/|\mathcal{R}|\rceil)+ W |\mathcal{R}|h(\lfloor\mu/|\mathcal{R}|\rfloor)$$
where $\mu$ is the total number of solutions, and $W$ is the total patient-day to be scheduled ($W = \sum_{p\in \mathcal{P}} |\mathcal{T}_p|$).

\section{Evolutionary Algorithm}\label{sec3}
We employ an EA to maximise the entropy of a set of PAS solutions (outlined in Algorithm \ref{alg:diversity_maximizing_EA}). The algorithm starts with $\mu$ copies of an optimal/near-optimal solution.  Having selected a parent uniformly at random, we generate an offspring by a mutation operator. If the quality of the offspring is acceptable, and it contributes to the entropy of the population $S$ more than the parent does, we replace the parent with the offspring; otherwise, it will be discarded. These steps are continued until a termination criterion is met. Since the PAS problem is heavily constrained, we use a low mutation rate in order to maintain the quality of solutions. This results in parents and offspring similar to each other, and there is no point to keep two similar solutions in the population while maximising diversity. Thus, the parent could be only replaced with its offspring.
\begin{algorithm}[H]
\begin{algorithmic}[1]
\REQUIRE{Initial population $S$, minimum quality threshold $c_{max}$}
\WHILE{The termination criterion is not met}
\STATE Choose $s \in S$ uniformly at random and generate one offspring $s'$ by the mutation operator\\ 
\IF{$O(s') \leq c_{max} \: \AND \: H(\{S \setminus s\} \cup s') > H(S)$}
\STATE Replace $s$ with $s'$
\ENDIF
\ENDWHILE
\end{algorithmic}
\caption{Diversity-Maximising-EA}
\label{alg:diversity_maximizing_EA}
\end{algorithm}
\subsection{Mutation}
In this sub-section, we introduce three mutation operators for the EA.
\subsubsection{Fixed change mutation:}
This mutation first selects $x$ patients uniformly at random and remove them from the solution. Let $\mathcal{P}_s$ denotes the set of selected patients. For each patient $p \in \mathcal{P}_s$, we identify those rooms that have enough capacity to accommodate them ($\mathcal{R}_{p}$). If no room has the capacity for patient $p$, we terminate the operator and return the parent. Otherwise, we calculate the cost of allocating patient $p$ to room $r \in \mathcal{R}_{p}$ ($C_{pr}$). Finally, we randomly allocate patient $p$ to one of the eligible rooms based on probability $pr(r, p)$ computed as follows: 
\begin{align*}
    pr(r, p) = \frac{C_{pr}^\gamma}{\sum_{r^{'}\in \mathcal{R}_{p}} C_{pr'}^\gamma}
    \hspace{1cm} \forall p \in \mathcal{P}_s; r \in \mathcal{R}_{p}
\end{align*}
where $\gamma$ is used to adjust the selection pressure. The above steps are repeated for all patients in $\mathcal{P}_s$. The fixed change mutation is outlined in Algorithm~\ref{chng}.\par 

Hyper-parameters $x$ and $\gamma$ should be passed to the operator, which will be tuned later for our problem. It is worth noting that we limit the search to the $y$ best rooms identified based on matrix $CV_{pr}$ for each patient to save computational cost. The standard swap, which has been used frequently in the PAS literature (e.g., see \cite{Cesc}), can be used as an alternative operator. In the standard swap, two patients are selected uniformly at random. Then, their rooms are swept for their whole LoS. While the standard swap might create infeasible offsprings, there is no such an issue for the fixed change mutation.

\begin{algorithm}[H]
\caption{Fixed change mutation}\label{chng}

\begin{algorithmic}[1]
\STATE Randomly select $x$ patients ($\mathcal{P}_s$)
\STATE Remove patients in $\mathcal{P}_s$ from allocated rooms
\FOR{$p$ in $\mathcal{P}_s$}
    \STATE Make a list of candidate rooms $R_{p}$ that they have the capacity for the LoS of patient $p$
    \STATE If $R_{p}$ is empty, terminate the operator and return the parent
    \FOR{$r$ in $R_{p}$}
        \STATE Evaluate the cost of allocating patient $p$ to room $r$ ($C_r$)
        \STATE Calculate $pr(r, p) = \frac{C_r^\gamma}{\sum_{r^{'}\in \mathcal{R}_{p}} C_{r^{'}}^\gamma}$
    \ENDFOR
    \STATE Randomly select a room $r$ using probability of $pr$
    \STATE Allocate patient $p$ to room $r$
\ENDFOR
\end{algorithmic}
\end{algorithm}

\subsubsection{Biased change mutation:}
Here, we make the fixed change mutation biased towards the entropy metric utilized for diversity maximisation. For a given room $r$, this new mutation operator selects patient $p \in \mathcal{P}_s$ according to probability $pr'(r, p)$.
\begin{align*}
    pr'(r, p) = \frac{\sum_{t\in \mathcal{T}_{p}} n_{{p}rt}}{\sum_{p^{'} \in \mathcal{P}_s}\sum_{t\in \mathcal{T}_{p^{'}}} n_{{p^{'}}rt}}
    \hspace{1cm} \forall p \in \mathcal{P}_s; r \in \mathcal{R}_{p}
\end{align*}

Obviously, the higher $pr'(r, p)$, the higher chance patient $p$ to be selected in the biased mutation. Thus, the most frequent assignments tend to occur less, which results in increasing the diversity metric $H(S)$.

\subsubsection{Self-adaptive change mutation:}
Adjusting hyper-parameters during the search procedure can boost efficiency of operators \cite{doerr2015optimal}. Based on initial experiments, we learned that the number of patients to be reallocated (i.e., hyper-parameter $x$) is the of great importance in the performance of both change and biased mutations, and can affect the performance of the EA more than the other hyper-parameters. Here, we employ a self-adaptive method to adjust parameter $x$ (similar to \cite{doerr2015optimal,neumann2017adapt}). As parameter $x$ increases, so does the difference between the parent and its offspring. However, a very large value for $x$ may result in poor quality offspring that do not meet the quality criterion. We assess $H(S)$ every $u$ fitness evaluations. If the algorithm has successfully increased $H(S)$, we raise $x$ to extend the changes in the offspring; otherwise, it usually indicates that offspring are low-quality and cannot contribute to the population. However, we limit $x$ to take values between $(x_{\min},x_{\max})$. In successful intervals, we reset $x = \max \{x \cdot F, x_{\min}\}$. And for failed intervals, $x = \min \{x \cdot F^{-\frac{1}{k}}, x_{\max}\}$. $F$ and $k$ here determine the adaptation scheme. The other hyper-parameters, such as $\gamma$, $k$, $u$, will be tuned by the iRace framework \cite{Lope}.

\section{Numerical Analysis}\label{sec4}
In this section, we tend to examine the performance of the EA using the existing benchmark instances. As mentioned earlier, there exist 13 test instances in the literature for the benchmark PAS problem. First, \cite{Deme1} introduced around half of test instances (Inst. 1-6), then, \cite{Deme2} introduced the other half (Inst. 7-13). Since test instances 1-6 have been further investigated in the literature, our study focuses on them. The specifications of these instances are reported in Table \ref{tab:spec}. 
\begin{table}[H]
    \centering
    \caption{Specifications of test instances: $B$, beds; $R$, rooms; $P$, patients; TP, total presence; PRC, average patient/room cost; BO, average percentage bed occupancy; SL, average LoS. \cite{Cesc}}
    \begin{tabular}{p{0.07\textwidth}|p{0.07\textwidth} p{0.07\textwidth} p{0.07\textwidth} p{0.07\textwidth} p{0.07\textwidth} p{0.07\textwidth} p{0.07\textwidth} p{0.07\textwidth}}
        \hline
        Inst. & $B$ & $R$ & $D$ & $P$ & TP & PRC & BO & SL \\
        \hline
        1 & 286 & 98 & 14 & 652 & 2390 & 32.16 & 59.69 & 3.66 \\
        2 & 465 & 151 & 14 & 755 & 3950 & 36.74 & 59.98 & 5.17 \\
        3 & 395 & 131 & 14 & 708 & 3156 & 35.96 & 57.07 & 4.46 \\
        4 & 471 & 155 & 14 & 746 & 3576 & 38.39 & 54.23 & 4.79 \\
        5 & 325 & 102 & 14 & 587 & 2244 & 31.23 & 49.32 & 3.82 \\
        6 & 313 & 104 & 14 & 685 & 2821 & 29.53 & 64.38 & 4.12 \\
        \hline
    \end{tabular}
    \label{tab:spec}
\end{table}

The diversity-based EA includes parameters whose values must be fully specified before use. The list of all parameters and their potential values for the fixed, adaptive and biased mutations are provided in Table \ref{tab:irace}. The choice of these parameters are significant for two main reasons. First, such algorithms are not parameter robust and might be inefficient with inappropriate parameter choices \cite{Nikf,Erfa}, and second, random choices of parameters would lead to an unfair comparison of the two algorithms \cite{Yuan}. For this problem, we apply the iterated racing package proposed by \cite{Lope} for automatic algorithm configuration (available on R as the iRace package). After specifying the termination criterion of the iRace package equal to 300 iterations, the best configurations of the EA are found and reported in Table \ref{tab:irace}. Note that we set the termination criterion for each run of the EA equal to $100,000$ fitness evaluations during the hyper-parameter tuning experiments due to limited computational budget. However, we set this parameter equal to $1,000,000$ fitness evaluations for the main numerical experiments (one fitness evaluation is equivalent to the computation of $H(S)$). Since $F$ and $k$ are two dependent parameter and together determine the adaptation scheme, we set $F$ equal to two based on initial experiments to make parameter tuning easier. Also, the population size is set to $50$ in the EA, and $\alpha \in \{0.02, 0.04, 0.16\}$. We consider 10 independent runs on each instances. \par
\begin{table}[H]
    \centering
    \caption{Parameter tuning for the EA}
    \begin{tabular}{l|l l |l l l l}
    \hline
        \multirow{2}{*}{Parameter} & \multirow{2}{*}{Type} & \multirow{2}{*}{Range} & \multicolumn{3}{l}{Elite configuration} \\
        \cline{4-7}
                                   &                       &                        & Fixed & Adaptive & Biased \\
    \hline
        $\gamma$    & Integer      & [15, 50]                       & 50 & 50  & 47  & \\
        $x$         & Integer      & [10, 30]                       & 14 & -   & -   & \\
        $x_{max}$   & Integer      & [10, 30]                       & -  & 15  & 14  & \\
        $k$         & Categorical  & \{0.25, 0.5, 1, 2, 4, 8, 16\}  & -  & 8   & 1   & \\
        $u$         & Categorical  & \{10, 50, 100, 200, 1000\}     & -  & 200 & 200 & \\
    \hline
    \end{tabular}
    \label{tab:irace}
\end{table}

\begin{figure*}
\centering
\includegraphics[width=.33\columnwidth]{In1Al2.pdf}
\hskip6pt
\includegraphics[width=.33\columnwidth]{In1Al16.pdf}
\includegraphics[width=.33\columnwidth]{In2Al2.pdf}
\hskip6pt
\includegraphics[width=.33\columnwidth]{In2Al16.pdf}
\begin{tikzpicture}
\node[opacity=.55] (legend) at (0,0) {\includegraphics[width=0.68\columnwidth,trim=5pt 10pt 2pt 0,clip]{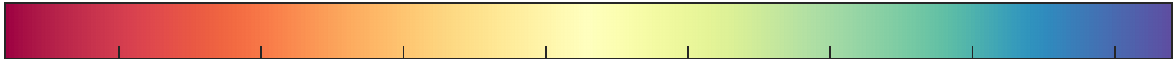}};
\node (5) at (-3.6,-0.3) {\scriptsize{\textcolor{gray!90}{low}}};
\node (15) at (0,-0.3) {\scriptsize{\textcolor{gray!90}{medium}}};
\node (25) at (3.6,-0.3) {\scriptsize{\textcolor{gray!90}{high}}};
\end{tikzpicture}
\caption{The distribution of the patients over rooms in the final populations obtained by the introduced EA on test instances $1$ and $2$ where $\alpha \in \{ 0.02, 0.16\}$.}
\label{fig:frq}
\end{figure*}
\vspace{-0.5cm}

The underlying aim of increasing $H(S)$ is to assign the patients to as many different rooms as possible. This increases the robustness of the solution population and provides decision-makers with more alternatives to choose. Figure \ref{fig:frq} illustrates the distribution of patients over rooms in the final populations obtained by the EA on test instance $1$ (first row) and $2$ (second row) where $\alpha = 0.02$ (first column) and $0.16$ (second column). Each cell represents a patients, and it is coloured based on the number of rooms allocated to them. Red cells are related to those patients with less changeable room assignments (given an acceptable cost threshold). On the other hand, blue cells associate with patients that the EA successfully assigned them to many different rooms. If $H(S) = 0$, all cells are coloured in red (like the initial population). Also, if $H(S) = H_{\max}$, the heat map only includes blue cells. Figure \ref{fig:frq} illustrates that the EA successfully diversifies the PAS solutions (several cells are not red), from which stakeholders have access to different alternatives with desirable quality. Also, these solutions can aid with imperfect modeling. For example, the cost of violating assumption $A2$ (i.e., the gender policy) is considered to be equal for all patients in the benchmark test instances. However, cost of such violations depend on several factors, including - but not limited to - patient's health condition, stakeholder preference and hospital policy. Having said that, it may be impossible to set an appropriate penalty for every single patient. The diverse set of solutions enables decision-makers to take health condition of patients into account. A patient would be assigned to a room in an inappropriate department in a high-quality (or even the optimal) solution to decrease the overall cost.\par
 \vspace{-0.5cm}
\begin{table}[H]
\centering
\caption{Comparison of the standard swap and three variants of the change mutation. Stat shows the results of Kruskal-Wallis statistical test at a $5\%$ significance level with Bonferroni correction. In row Stat, the notation $X^+$ means the median of the measure is better than the one for variant $X$, $X^-$ means it is worse, and $X^*$ indicates no significant difference.}
\renewcommand{\tabcolsep}{2pt}
\renewcommand{\arraystretch}{0.7}
\begin{tabular}{ll|cc|cc|cc|cc|c}
\toprule
        & &Swap&&Fixed&Change&Adaptive&Change&Biased&Change&\multirow{3}{*}{$H_{max}$}\\
\cmidrule(l{2pt}r{2pt}){3-4}
\cmidrule(l{2pt}r{2pt}){5-6}
\cmidrule(l{2pt}r{2pt}){7-8}
\cmidrule(l{2pt}r{2pt}){9-10}

             $\alpha$  & Inst. & H(S) & Stat (1) & H(S) & Stat (2) & H(S) & Stat (3) & H(S) & Stat (4) \\
\midrule
2&1&4865.8&$2^-3^-4^*$&6746.3&$1^+3^*4^*$&\hl{6752.4}&$1^+2^*4^+$&6679.1&$1^*2^*3^-$ & \multirow{3}{*}{13488.8}\\
4&1&4966.3&$2^-3^-4^*$&6895.9&$1^+3^*4^*$&\hl{6920.9}&$1^+2^*4^+$&6818.9&$1^*2^*3^-$\\
16&1&5426.7&$2^-3^-4^*$&7660&$1^+3^*4^*$&\hl{7691.4}&$1^+2^*4^+$&7500.1&$1^*2^*3^-$\\
\midrule
2&2&7734.2&$2^-3^-4^*$&\hl{11563.3}&$1^+3^*4^+$&11559.5&$1^+2^*4^+$&11374.1&$1^*2^-3^-$&\multirow{3}{*}{22039.2}\\
4&2&7889.4&$2^-3^-4^*$&\hl{11971.9}&$1^+3^*4^*$&11957.3&$1^+2^*4^+$&11718.1&$1^*2^*3^-$\\
16&2&8633.4&$2^-3^-4^*$&13597.5&$1^+3^*4^*$&\hl{13633.2}&$1^+2^*4^+$&13448.1&$1^*2^*3^-$\\
\midrule
2&3&6000.6&$2^-3^-4^*$&9043.6&$1^+3^*4^*$&\hl{9061.5}&$1^+2^*4^+$&9011.3&$1^*2^*3^-$&\multirow{3}{*}{17812}\\
4&3&6128&$2^-3^-4^*$&9332.7&$1^+3^*4^*$&\hl{9353.6}&$1^+2^*4^+$&9242&$1^*2^*3^-$\\
16&3&6706.5&$2^-3^-4^*$&10499.6&$1^+3^*4^+$&\hl{10503.3}&$1^+2^*4^+$&10381.8&$1^*2^-3^-$\\
\midrule
2&4&6849.9&$2^-3^-4^*$&10273.3&$1^+3^*4^*$&\hl{10278.9}&$1^+2^*4^+$&10225.2&$1^*2^*3^-$&\multirow{3}{*}{20812.4}\\
4&4&6986.5&$2^-3^-4^*$&10590.5&$1^+3^*4^*$&\hl{10620.2}&$1^+2^*4^+$&10507.5&$1^*2^*3^-$\\
16&4&6864.3&$2^-3^-4^*$&11989&$1^+3^*4^*$&\hl{12016.9}&$1^+2^*4^+$&11881.1&$1^*2^*3^-$\\
\midrule
2&5&6446.5&$2^-3^-4^-$&7772&$1^+3^*4^-$&7774.9&$1^+2^*4^-$&\hl{7835.6}&$1^+2^+3^+$&\multirow{3}{*}{12664.8}\\
4&5&6542.3&$2^*3^-4^-$&8081.4&$1^*3^*4^-$&8090.9&$1^+2^*4^*$&\hl{8142.7}&$1^+2^+3^*$\\
16&5&7012.7&$2^*3^-4^-$&7775.3&$1^*3^*4^-$&8965.8&$1^+2^*4^*$&\hl{9069.3}&$1^+2^+3^*$\\
\midrule
2&6&6163.6&$2^-3^-4^*$&8368.2&$1^+3^*4^*$&\hl{8381.9}&$1^+2^*4^+$&8328.5&$1^*2^*3^-$&\multirow{3}{*}{15921.3}\\
4&6&6293.1&$2^-3^-4^*$&8671.5&$1^+3^*4^*$&\hl{8680.9}&$1^+2^*4^+$&8574.5&$1^*2^*3^-$\\
16&6&6905.7&$2^-3^-4^*$&9857.8&$1^+3^*4^*$&\hl{9865.1}&$1^+2^*4^+$&9772.9&$1^*2^*3^-$\\

\bottomrule
\end{tabular}
\label{tbl:Res_swap}
\end{table}
We now compare the standard swap and three variants of the change mutation. Table \ref{tbl:Res_swap} summarises the results of the EA obtained using the swap (1), and fixed, biased and self-adaptive change mutations (2-4). In this table, $H_{\max}$ presents the maximum entropy where no quality constraint is imposed. It is highly unlikely to achieve $H_{\max}$ when the quality constraint is considered, but we use it as an upper-bound for $H(S)$. The table shows the superiority of all variants of the change mutation over the standard swap over all six instances. In fact, the change mutations result in $31\%$ higher entropy on average compared to the standard swap. The self-adaptive mutation provides the average highest $H(S)$ over instances 1, 3, 4, 6. This is while the fixed mutation and the biased have the best results for instances 2 and 5, respectively. We can observe the same results for all $\alpha \in \{0.02, 0.04, 0.16\}$. All aforementioned observations are confirmed by the Kruskal-Wallis statistical test at a $5\%$ significance level and with Bonferroni correction.



\begin{figure}[H]
\centering
\includegraphics[width=0.9\columnwidth]{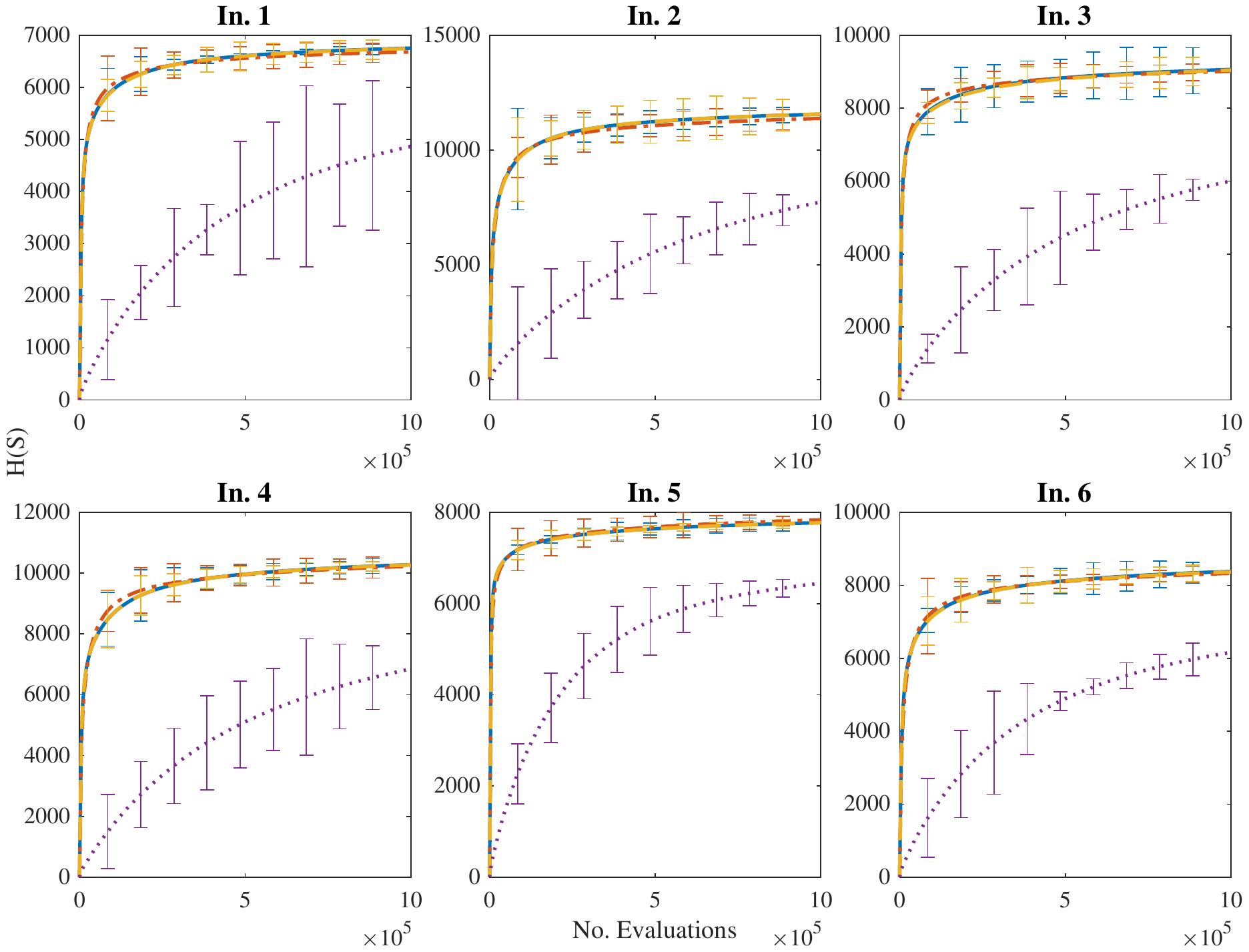}
\includegraphics[width=0.8\columnwidth]{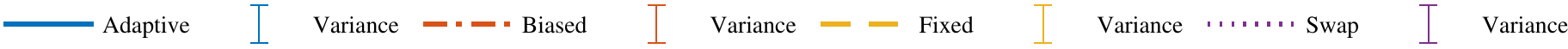}

\caption{Representation of trajectories of the EA employing different operators ($\alpha = 2\%$).}
\label{fig:Trd}
\end{figure}

Figure \ref{fig:Trd} illustrates the trajectories of the average entropy of the population obtained by the EA using different operators over 10 runs. This figure indicates that all variants of the change mutation converge faster and to a higher value of $H(S)$ than the standard swap across all test instances. It also shows that the majority of improvements for the change mutations have occurred in the first $100,000$ fitness evaluations while the gradient is slower for the swap. Among the change mutations, the biased one has the highest $H(S)$ in the first $100,000$ fitness evaluations, but it gets overtaken by the self-adaptive mutation during the search over all instances except instance 5. The same observation (with smaller gap) can be seen between the fixed and self-adaptive mutations, where the fixed mutation is overtaken by the self-adaptive one during the last $200,000$ fitness evaluations except instance 2. Since instance 2 is the largest instance, the self-adaptive mutation may cross the fixed one if the EAs run for a larger number of fitness evaluations. Also, the figure depicts the variance of the entropy of populations obtained by the standard swap is considerably higher than those obtained by the fixed, adaptive, and biased change mutations.

Earlier, we mentioned how the EA can provide decision-makers with robustness against imperfect modeling. To evaluate this robustness for population $S$, for instance, we investigate a scenario that pairs of patients should not share the same room. For this purpose, we randomly choose $b \in \{1, 4, 7\}$ pairs of patients that accommodate the same rooms in the initial population/solution. Repeating this selection for 100 times (e.g., the number of selected pairs is equal to $4 \times 100$ if $b=4$), we then assess: (i) the ratio of times that all patient pairs are allocated to different rooms in at least one of the solutions in the population (Ratio), and (ii) the average number of solutions that all patient pairs occupy different rooms across 100 selections (Alt). Table \ref{tbl:Res_sim} presents the results for this analysis. As shown in the table, the ratio is always higher than $95\%$. It also indicates that $S$ includes more than $20$ alternatives when $\alpha=2\%$ and $b=7$ (the worst case). This indicates that the proposed EA can successfully diversify $S$ and robust against imperfect modeling.

\begin{table}[H]
\centering\caption{Results for the analysis of the EA in terms of the robustness}
\renewcommand{\tabcolsep}{0.7pt}
\renewcommand{\arraystretch}{0.9}
\begin{tabular}{l|cccccc|cccccc|cccccc}
\toprule
    b&&&1&&&&&&4&&&&&&7&&&\\
\cmidrule(l{2pt}r{2pt}){2-7}
\cmidrule(l{2pt}r{2pt}){8-13}
\cmidrule(l{2pt}r{2pt}){14-19}
             $\alpha$&0.02& &0.04&
             &0.16 &&0.02&&0.04&
             &0.16 &&0.02&&0.04&
             &0.16& \\
\cmidrule(l{2pt}r{2pt}){2-3}
\cmidrule(l{2pt}r{2pt}){4-5}
\cmidrule(l{2pt}r{2pt}){6-7}
\cmidrule(l{2pt}r{2pt}){8-9}
\cmidrule(l{2pt}r{2pt}){10-11}
\cmidrule(l{2pt}r{2pt}){12-13}
\cmidrule(l{2pt}r{2pt}){14-15}
\cmidrule(l{2pt}r{2pt}){16-17}
\cmidrule(l{2pt}r{2pt}){18-19}
            Inst. & Ratio  & Alt & Ratio  & Alt& Ratio  & Alt& Ratio  & Alt& Ratio  & Alt& Ratio  & Alt& Ratio  & Alt& Ratio  & Alt& Ratio  & Alt\\
\midrule
1&99&45.4&100&43.9&100&45.8&99&32.8&100&31.4&100&38.2&96&21.1&96&21.4&100&28.2\\
2&100&44.7&100&45.9&100&46.6&100&32&100&31.4&100&36.7&99&21.8&99&24&100&30.1\\
3&99&43.9&100&45&100&46.1&97&29.9&99&31.6&100&36.3&98&22.5&99&23&100&29.3\\
4&100&44&100&42.7&100&45.5&100&30.6&100&30.8&100&36.7&99&22.1&99&22.4&100&27.8\\
5&100&46.1&100&47&100&47.8&100&33.7&100&36.5&100&42.1&99&25.6&100&28.9&100&38\\
6&99&44.7&99&44.9&100&46.9&98&34.6&99&35.5&100&38.2&96&24&96&27.1&100&33.9\\
\bottomrule
\end{tabular}
\label{tbl:Res_sim}
\end{table}

\section{Conclusions and Remarks}\label{sec5}
In this study, we introduced a methodology to compute a highly diverse set of solutions for the PAS problem. We first defined an entropy-based measure to quantify the diversity of PAS solutions. Then, we proposed an EA to maximise the diversity of solutions and introduced three mutations for the problem. The iRace package was used to tune the hyper-parameters of the EA. Through a comprehensive numerical analysis, we demonstrated the efficiency of the proposed mutations in comparison to the standard swap. Also, our analyses revealed that the EA is capable of computing high-quality and diverse sets of solutions for the PAS problem. Finally, we showed the robustness of solutions found by the proposed algorithm against imperfect modeling by performing a scenario analysis.

For future studies, it could be interesting to investigate the assumption that optimal/near-optimal solutions are not known a-priori for the PAS problem. It might be also valuable to apply diversity-based EAs to other real-world optimisation problems in healthcare, such as the operating room planning and scheduling, home-health care routing and scheduling, and nurse scheduling. These problems are usually multi-stakeholder (with conflicting interests) and require high robust solutions.  

\section*{Acknowledgements}
This work was supported by the Australian Research Council through grants DP190103894 and FT200100536.

\bibliographystyle{splncs04}
\bibliography{Main}
\end{document}